\documentclass[conference,10pt,letter]{IEEEtran}
\IEEEoverridecommandlockouts
\usepackage{cite}
\usepackage{amsmath,amssymb,amsfonts}

\usepackage{textcomp}
\usepackage[table]{xcolor}
\usepackage{wrapfig}
\usepackage{subfig}
\usepackage[font=small,labelfont=small]{caption}
\usepackage[linesnumbered,ruled]{algorithm2e}%\RestyleAlgo{ruled}
\usepackage{xpatch}
\usepackage{multirow}
\usepackage{multicol}
\usepackage{wrapfig}
\usepackage[export]{adjustbox}
\usepackage[numbers,sort&compress]{natbib}
\usepackage{nomencl}
\usepackage{siunitx}
\usepackage{blindtext}
\usepackage[T1]{fontenc}
\usepackage[spaces,hyphens]{xurl}
\usepackage{float}
\usepackage{enumitem}
\usepackage{graphicx}
\usepackage[
            left=0.52in,
            right=0.52in,
            top=0.58in,
            bottom=0.33in,
            ]{geometry}
\SetAlFnt{\small}

\SetKwComment{Comment}{$\triangleright$\ }{}

\usepackage[utf8]{inputenc}

\SetKwInput{kwInit}{Initialize}

\usepackage{fancyhdr}
\fancypagestyle{mystyle}{
    \chead{2023 IEEE International Conference on Communications (ICC 2023)}
}

\def\BibTeX{{\rm B\kern-.05em{\sc i\kern-.025em b}\kern-.08em
    T\kern-.1667em\lower.7ex\hbox{E}\kern-.125emX}}

\usepackage[numbers,sort&compress]{natbib}

\newcommand{\fref}[1]{Fig.~\ref{#1}}

\newcommand{\sref}[1]{Section~\ref{#1}}

\newcommand*{\Resize}[2]{\resizebox{#1}{!}{$#2$}}%

\newcommand\HUGE{\fontsize{20.8}{25}\selectfont}
\begin{document}
\title{\HUGE \vspace{0.1cm} Optimizing Federated Learning in LEO Satellite Constellations via Intra-Plane Model Propagation and Sink Satellite Scheduling\vspace{-0.4cm}}
\author{Mohamed Elmahallawy, Tie Luo$^*$\thanks{$^*$Corresponding author.\vspace*{14mm}}\\ {Computer Science Department, Missouri University of Science and Technology, USA}\\ {E-mail: \{meqxk, tluo\}}@mst.edu \vspace*{-0.5cm}}
\maketitle
\thispagestyle{mystyle}

\begin{abstract}

The advances in satellite technology developments have recently seen a large number of small satellites being launched into space on Low Earth orbit (LEO) to collect massive data such as Earth observational imagery. The traditional way which downloads such data to a ground station (GS) to train a machine learning (ML) model is not desirable due to the bandwidth limitation and intermittent connectivity between LEO satellites and the GS. Satellite edge computing (SEC), on the other hand, allows each satellite to train an ML model onboard and uploads only the model to the GS which appears to be a promising concept. This paper proposes {\em FedLEO}, a novel federated learning (FL) framework that realizes the concept of SEC and overcomes the limitation (slow convergence) of existing FL-based solutions. FedLEO (1) augments the conventional FL's star topology with ``horizontal'' intra-plane communication pathways in which {\em model propagation} among satellites takes place; (2) optimally schedules communication between ``sink'' satellites and the GS by exploiting the {\em predictability} of satellite orbiting patterns. We evaluate FedLEO extensively and benchmark it with the state of the art. Our results show that FedLEO drastically expedites FL convergence, without sacrificing---in fact it considerably increases---the model accuracy.

\end{abstract}
\vspace{-1mm}
\section{Introduction}\label{sec:intro}

With the fast developments of satellite technology, a large number of low Earth orbit (LEO) satellites have been launched into space. Among the active players in LEO satellite deployment are government agencies such as NASA and ESA, and large companies such as DigitalGlobe, SpaceX (Starlink), and OneWeb. They offer 1) Earth observational and monitoring services or 2) broadband Internet services. For the first category which this paper focuses on, a massive amount of data such as high-resolution images are continuously collected by LEO satellites. Performing data analytics on such big data using AI techniques in order to derive insights and inform decision-makers has huge value and implications, for example in disaster early warning \cite{barmpoutis2020review} and urban planning \cite{happaper}. In the traditional school of thought, we can download all these images to a central server such as a ground station (GS), and then train some complex machine learning (ML) models to perform data analytics. However, this is not desirable in satellite communication (Satcom) because of 1) the network bandwidth limitation \cite{happaper}, 2) the large propagation delay between satellites and the GS \cite{razmi}, 3) the short and irregular visibility period between satellites and the GS \cite{mAsyFLEO}, and 4) the need for low-latency tracking of space missions \cite{vasisht2021l2d2}. 

Federated Learning (FL) \cite{mcmahan2017communication} offers a promising solution to this problem. Instead of downloading raw data, FL realizes the concept of satellite edge computing (SEC) by enabling each satellite to train a local ML model onboard and upload only the model parameters to the GS to aggregate into a global model. Although this solves the aforementioned bandwidth issue, the learning process of FL will incur a large delay in Satcom because FL is an iterative process that requires many communication rounds between clients (LEO satellites) and the central server (GS) for training. In Satcom, LEO satellites and the GS move in distinct directions\footnote{A GS always rotates on the 0$^o$ plane (in reference to the Equator), while a satellite flies in an orbit of inclination angle between 0-90$^o$. Thus, there are very irregular and large delays between successive visits of a satellite to the same GS.\vspace*{14mm}} and speeds, which leads to highly {\em intermittent and irregular visibility} between them, and thus each communication round takes significantly longer and the entire training process takes several days or even weeks to converge \cite{so2022fedspace}.

In this paper, we propose FedLEO, a novel FL framework for LEO satellite constellations that accelerates the FL training process in Satcom. FedLEO consists of two novel components:
\begin{enumerate}[label=\arabic*)]%[leftmargin=*]
\item We propose a {\em model propagation} scheme that augments the conventional ``vertical'' star topology between the FL server and LEO satellites with ``horizontal'' intra-plane communication pathways among satellites;

\item We propose a {\em distributed scheduling} algorithm that selects an optimal ``sink'' satellite on each orbit to generate a {\em partial global model}, and optimally schedules communication between sink satellites and the GS by exploiting the {\em predictability} of satellite orbiting patterns.
%on each orbit that will have the soonest and longest visible period to the GS to exchange models with the GS.
\end{enumerate}

In addition, this paper also makes the following contributions:
\begin{enumerate}\addtocounter{enumi}{2}%[leftmargin=*]
\item FedLEO addresses the challenging issue of {data discrepancy} between satellites on different orbits by taking care of non-independent and identically distributed (non-IID) data.

\item We demonstrate via extensive simulations that FedLEO drastically accelerates FL convergence by {\em more than an order of magnitude} than state-of-the-art FL-Satcom methods. Moreover, despite the much shorter time, FedLEO achieves much higher accuracy than those benchmarks.

\end{enumerate}

%To the best of our knowledge, FedLEO is the first FL-Satcom framework leveraging on-orbit collaboration among LEO satellites to address the challenge of highly intermittent connectivity between satellites and GS without sacrificing any desirable FL requirements (e.g., accuracy and convergence time) or requiring the download of any raw data in any format to the GS.

\vspace*{-1mm}
\section{Related Work} \label{sec:releated}

In general, FL can be categorized into synchronous and asynchronous approaches. We discuss these two categories in the context of Satcom.

In synchronous FL, the central server must wait to receive all the clients' models and then aggregate them into a global model. As a result, slow clients, often referred to as ``stragglers'', will become the bottleneck of the training process. A few recent studies have attempted this synchronous approach in Satcom. For example, \cite{chen2022satellite} simulated the vanilla FL (i.e, FedAvg \cite{mcmahan2017communication}) in LEO constellations and show that FL is more advantageous than directly downloading raw data to GS to perform centralized training. Razmi et al. proposed FedISL \cite{razmi} to leverage inter-satellite links (ISL) to reduce delay. They use an ideal setup in which the server is either a GS located at the North Pole (NP) or a medium Earth orbit (MEO) satellite located above the Equator (at an altitude of 20,000 km), so that each satellite visits the  server at regular intervals (i.e., no longer irregularly) and much more often. This ideal setup is rarely available in practice. A most recent approach called FedHAP \cite{happaper} was proposed which substitutes the traditional GS by one/multiple high altitude platforms (HAPs) floating at 20-30 km above the Earth's surface to act as servers. Although the performance improves, it requires extra hardware.

On the other hand, an asynchronous FL approach allows the server to proceed to the next training round with just a {\em subset} of the clients' model updates received, instead of waiting for all. This mitigates the bottleneck problem existing in synchronous FL, but it faces a {\em staleness} problem where some received model updates in a certain round may come from earlier rounds (due to stragglers). FedSat \cite{razmi2022ground} is such an asynchronous FL approach applied to Satcom, and it again assumes the above ideal setup where the GS is located at NP to avoid the challenge of irregular satellite visit patterns. Another work \cite{razmischeduling} attempts to reduce model staleness by calculating whether the visible interval of each satellite is sufficient for global model downloading, local model training, and uploading; if not, it will schedule local model uploading to the next communication round so that the satellite can train local model during the (long) invisible interval. %by predicting satellites' visiting patterns to the GS 
%\red{This sounds too similar to what you are doing. What's the difference?}\blue{this scheduler only determines whether or not the on-time (visible time) is enough for model exchange. If so, then the PS transmits the global model to SAT and waits until it receives back its local model. If not, then the PS will transmit the global model only to the SAT and schedule to receive its local model to the next comm round. So, it  allows this SAT to train its model within the off-time period (invisible time). Also, this paper offered an async FL approach without considering the ISL}. 
However, this has only a limited effect on efficiency, which still requires several days to finish training a global model. So et al. proposed another asynchronous FL algorithm called FedSpace \cite{so2022fedspace}, which collects client models into a buffer with a predicted size, and down-weights stale models. However, it requires each satellite to upload a small portion of local data to the GS for aggregation scheduling, which violates the FL principle of privacy protection by avoiding raw data sharing. Recently, we proposed AsyncFLEO \cite{mAsyFLEO}, which tackles the model staleness issue and achieves fast convergence within a few hours. However, it still requires HAPs to serve as servers like FedHAP \cite{happaper}. %Despite this progress, there remains an unresolved challenge of ensuring sufficient visibility time for each sink satellite to train and exchange models with the server, offering the potential for further improvement. 
Different from all the above, FedLEO does not assume ideal setups, violate any FL principle, or introduce any extra hardware (e.g., HAP). While it is a synchronous protocol, FedLEO achieves the lowest FL convergence delay in Satcom yet improves accuracy at the same time, in a realistic setting including irregular satellite visiting patterns, bandwidth limitation, and non-IID data. 
\vspace{-3mm}\label{sec:model}\enlargethispage{-3.7\baselineskip}
\section{System Model} 

\begin{figure}[t]
    \vspace{0.2cm}
     \centering
     \includegraphics[width=0.8\linewidth]{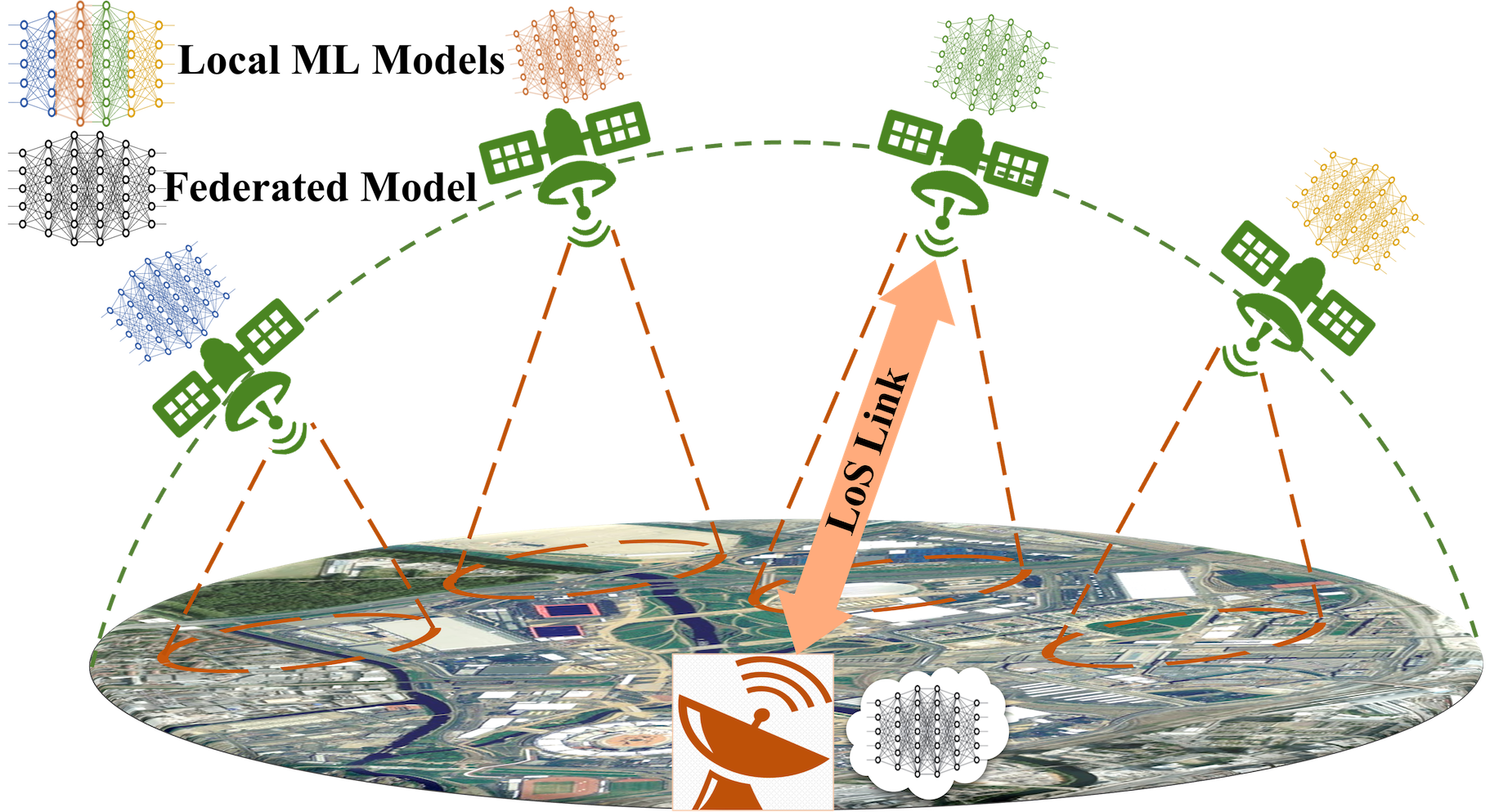}
    \caption{FL in the context of an LEO satellite constellation, with a GS serving as the federation server. Only one orbit is shown for clarity.}
    \label{Earth_observation}
    \vspace{-0.3cm}
\end{figure}

%FedLEO can be applied to any generic LEO satellite constellation. %at any orbital period with different inclinations. 
As illustrated in \fref{Earth_observation}, we consider a generic LEO satellite constellation $\mathcal K$ consisting of $L$ orbits with each having $K$ equally distributed satellites. An orbit $l \in \mathcal{L}=\{l_1,l_2,...,l_L\}$ contains its own set of satellites $\mathcal{K}_{l}$, and is characterized by its inclination angle $\alpha_{l}$ and altitude $h_{l}$. Each satellite in orbit $l$ travels at the same speed $v_{l}$ and has the same orbital period $T_{l}$. Here, $v_{l}=\sqrt{\frac{GM}{(R_{E}+h_{l})}}$ and $T_{l}= \frac{2 \pi}{\sqrt{GM}}{(R_{E}+h_{l})^{3/2}}$, where $G$ denotes the gravitational constant; $M$ denotes the mass of the Earth; and $R_{E}=6371$ km is the radius of the Earth. Satellites and the GS typically use a radio frequency (RF) link that is more reliable than free-space optical (FSO) for long-distance communication. Thus, any satellite $k\in \mathcal K$ will be visible to a GS $g$ when the line of sight (LoS) link is not blocked by the Earth. This can be translated mathematically to the condition $\vartheta_{k,g}(t) \triangleq \angle (r_{g}(t), (r_{k}(t) - r_{g}(t))) \leq \frac{\pi}{2}-\vartheta_{min}$, where $r_{k}(t)$ and $r_{g}(t)$ represent the trajectory of a satellite \textit{k} and a GS \textit{g}, respectively, and $\vartheta_{min}$ represents the minimum elevation angle, a constant depending on the GS position. 

\vspace{-1mm}
\subsection{Computation Model}

Consider an FL-LEO system that collects data such as Earth observational imagery. Each satellite $k$ in the constellation $\mathcal K$ captures a set of images, which we assume to be non-IID due to their different orbits and positions. Let $\mathcal {D}_{k}=\{(X_{k,1}, y_{k,1}), (X_{k,2}, y_{k,2}), \dots, ((X_{k,m_{k}}, y_{k,m_{k}})\}$ represents the local dataset collected by satellite $k$, where $X_{k,j}$ denotes the feature vector of the \textit{j-th} sample, and $y_{k,j}$ denotes its corresponding label (for notation purposes only; labels are not required), and $m_{k}=|\mathcal {D}_{k}|$ is the dataset size. The objective of the FL system is to train an ML model by the GS  and the satellites collaboratively; specifically, to solve\vspace{-0.1cm}
\begin{equation} \label{eqn1}
%    \begin{array}{rrclcl} \displaystyle 
           \arg \min_{w \in \mathbb{R}^{d}} F(w)= \sum_{k\in \mathcal K }{\frac{m_{k}}{m}{F_{k}{(w)}},}
%    \end{array}
\vspace{-0.15cm}
\end{equation}
where $F(w)$ is the overall loss function of the model; $w$ denotes the model weights; and $m= \sum_{i\in \mathcal K} m_{k}$ is the total data size. $F_{k}(w)$ below is the loss function of satellite $k$
\begin{equation}\vspace{-0.1cm}
     F_{k}{(w)} = \frac{1}{m_{k}}\sum_{j=1}^{m_{k}} f_{k}(w; X_{k,j}, y_{k,j}),
     \vspace{-0.1cm}
\end{equation}
where $f_{k}(\cdot)$ is the training loss on a sample $X_{k,j}$.

In a sync FL system such as FedAvg \cite{mcmahan2017communication}, the training process takes multiple communication rounds $t=0, 1, 2, \dots, T$.  In each round, the GS broadcasts the global model $w^t$ to all satellites that are in its visible zone. Each satellite $k$ then applies a local optimization method such as stochastic gradient descent (SGD) for $I$ local epochs as  
\begin{equation}
    w_{k}^{t,i+1} = w_{k}^{t,i}- \eta \nabla F_{k}(w_{k}^{t,i}; X_{k}^{i}, y_{k}^{i}), ~ i=0,1,2,...,I-1
\end{equation}
where $\eta$ is the learning rate. Then, the satellites transmit their updated model weights (or gradients) to the GS, when entering its visible zone. Once the GS receives all satellite models from all the orbits, it lumps them into a global model as follows:
\begin{equation}
    w^{t+1} = \sum_{k\in \mathcal K} \frac{m_{k}}{m} w_{k}^{t,I},
\end{equation} \enlargethispage{-3.5\baselineskip}
The above procedure iterates with an incremental $t$ until the FL model converges (e.g., a target loss, target accuracy, or the maximum number of training rounds $T$ is reached).

A key problem with this learning process is that all the communications (upload/download) can only happen when a satellite transiently comes into the GS's visible zone, which causes the entire process to take days or even weeks. This also implies that the learning speed is not able to catch up with the data generation or collection speed. As a result, the model will always be out-dated.

%our FedLEO framework proposes two novel algorithms to deal with these challenges: the propagation algorithm and the scheduling algorithm.
\vspace*{-1mm}
\subsection{Communication Model}\label{Com_link}
The signal-to-noise ratio (SNR) between any satellite $k$ and the GS on a symmetric RF channel with additive white Gaussian noise (AWGN) can be defined as \cite{mAsyFLEO}
\begin{equation}
    SNR(k,GS) = \frac{P_{t}G_{k}G_{GS}}{K_{B} T B {L}_{k,GS}}, 
\end{equation}
where $P_{t}$ is the sender power, $G_{k}$ and $G_{GS}$ are the total antenna gains of the \textit{k-th} satellite and the GS, respectively, $K_{B}$ is the Boltzmann constant, %($1.380649 10^-23 J/K$), 
\textit{T} is the noise temperature at the receiver, \textit{B} is the channel bandwidth, and ${L}_{k,GS}$ denotes the free-space pass loss between a satellite $k$ and the GS. As long as the LoS link between the satellite $k$ and the GS is established (i.e., Earth does not block the communication between them), then we can express ${L}_{k,GS}$ as follows:
\begin{equation}
    {L}_{k,GS} = \bigg(\frac{4\pi \|k,GS\|_{2}  f}{c}\bigg)^{2}  
\end{equation}
where $\|k,GS\|_{2}$ is the Euclidean distance between the \textit{k-th} satellite and the GS if LoS is achieved, $f$ is the carrier frequency, and \textit{c} is the speed of the light. For exchanging local or global model weights ($w_k~\text{or}~w$) between a satellite $k$ and the GS, the total required time $t_{c}$ can be calculated as
\begin{align}
    t_{c} &= t_{t}+t_{p}+t_{k}+ t_{GS}, ~~
    t_{t}=\frac{z{|\bf{\mathcal{N}|}}}{R} , ~~ t_{p} = \frac{\|k,GS\|_{2}}{c}, \label{eqny}
\end{align}
where $t_t$ and $t_p$ are the transmission and propagation times, respectively, $t_{k}$ and $t_{GS}$ are the processing delay at \textit{k-th} satellite and the GS, respectively (we omit them in our simulation since they are much smaller than $t_t$ and $t_p$), ${|\mathcal{N}|}$ is the number of data samples, $z$ is the number of bits in each sample, and $R$ is the maximum achievable data rate, which can be approximated by the Shannon formula as
\begin{equation} \label{eqnz}
    R \approx B \log_{2} (1+SNR(k,GS))
\end{equation}

\vspace*{-2mm}
\section{FedLEO Framework}
FedLEO consists of two main components, model propagation, and distributed scheduling, which we describe below.
%i) {\em propagating algorithm} that leverages intra-plane ISL (inter-satellite link) to facilitate satellite collaboration for propagating local/global models amongst them, thereby addressing the challenge of intermittent connectivity, and (ii) {\em scheduling algorithm} for selecting a {\em sink satellite} that has the most suitable visible interval with the GS, thereby avoiding Earth-blockage of LoS between LEO satellites and the GS, as well as avoiding satellites with limited visibility periods.
\subsection{Model Propagation} \label{propagate}

\begin{figure*}[t]
\vspace{0.1cm}
\centering
          \subfloat[Broadcasting global model $w^t$ to all visible satellites. 3 orbits are differentiated by 3 colors.]
         {\centering
         \includegraphics[width=0.32\textwidth]{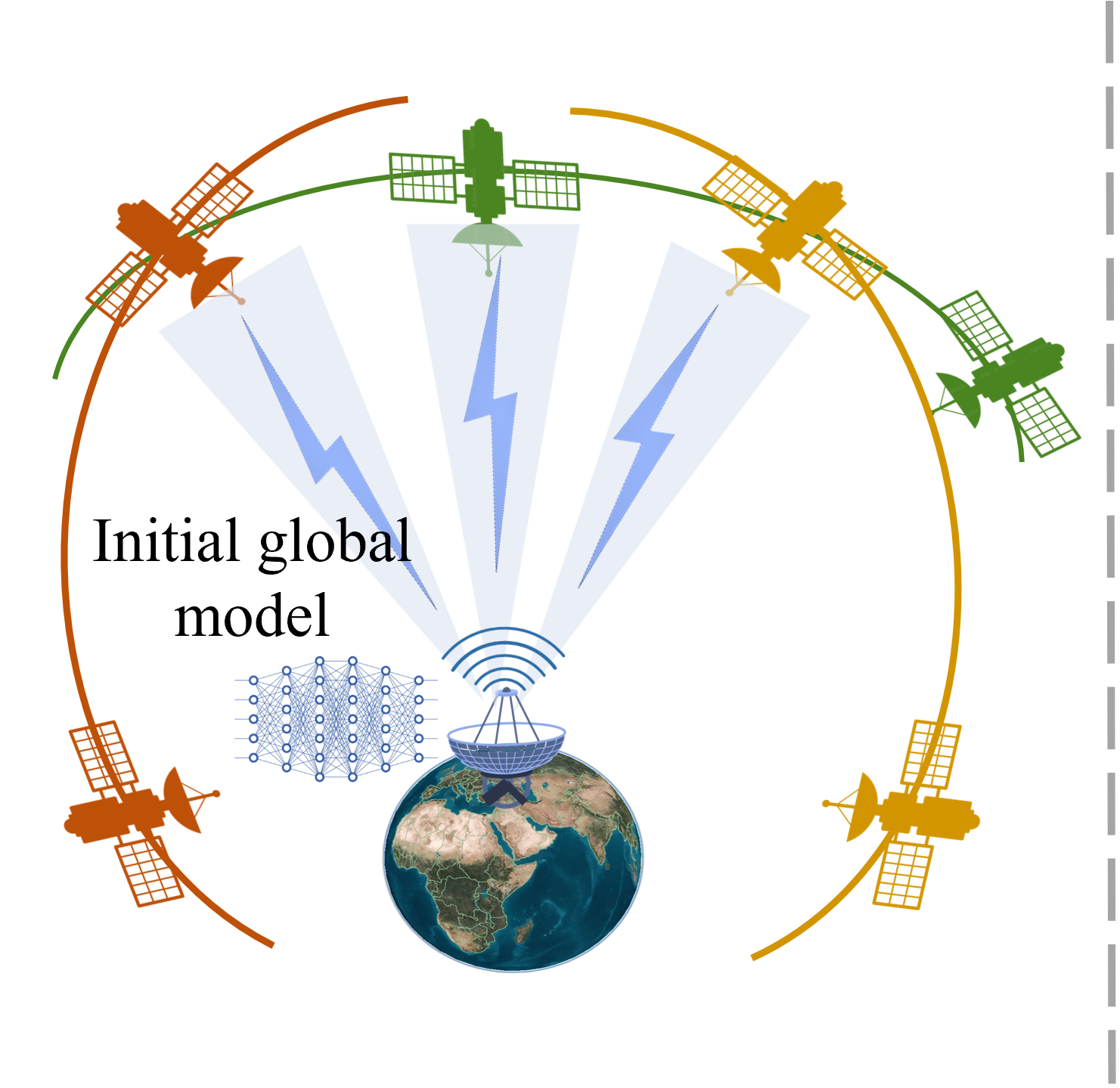}}
     \hfill
          \subfloat[Propagating a model ($w^t$) from visible satellite (top) to all other satellites on the same orbit.]
         {\includegraphics[width=0.33\textwidth]{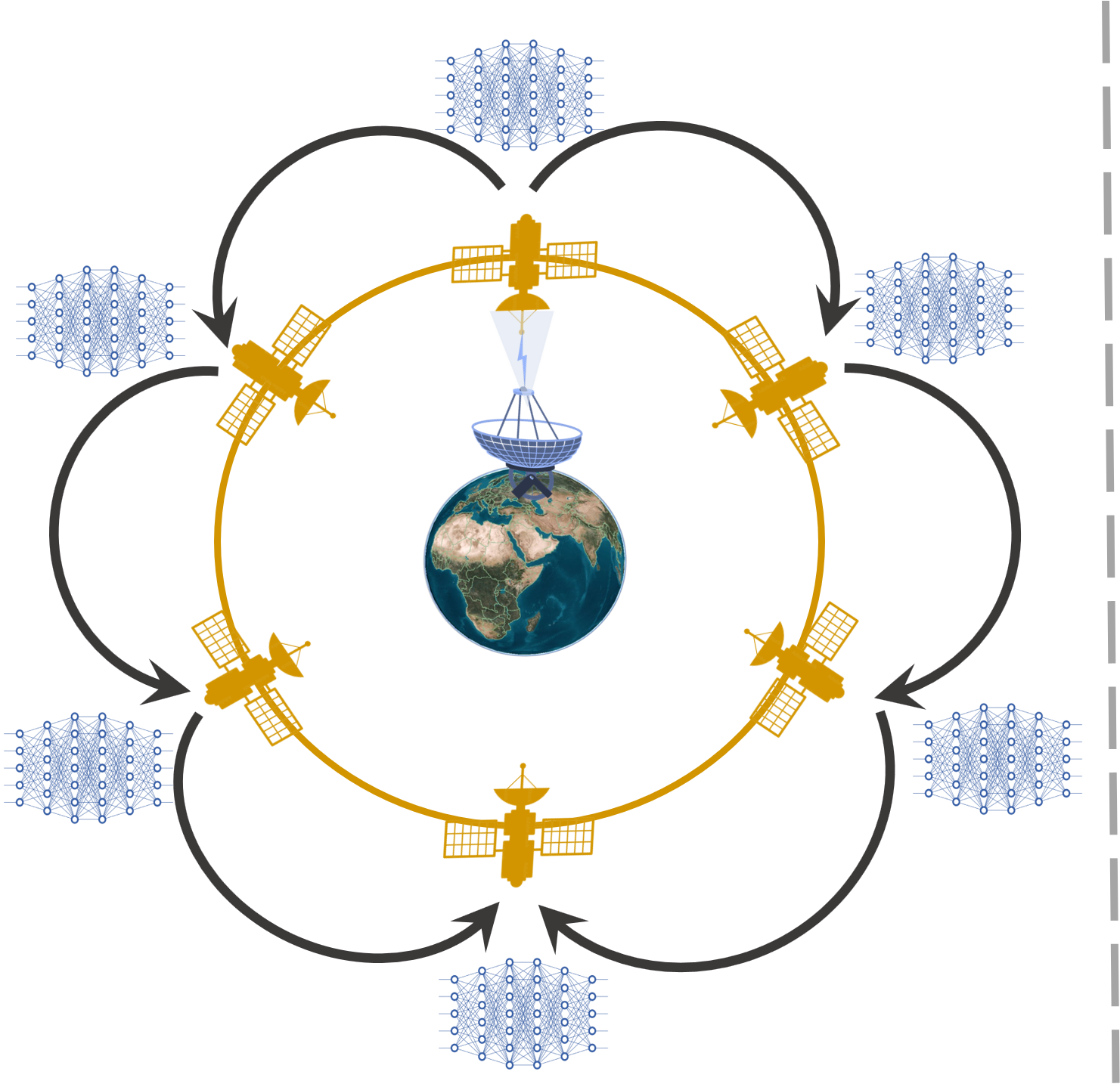}}
     \hfill
       \subfloat[After local training, relaying all the trained models to the sink satellite (in blue).]
      {\includegraphics[width=0.3\textwidth]{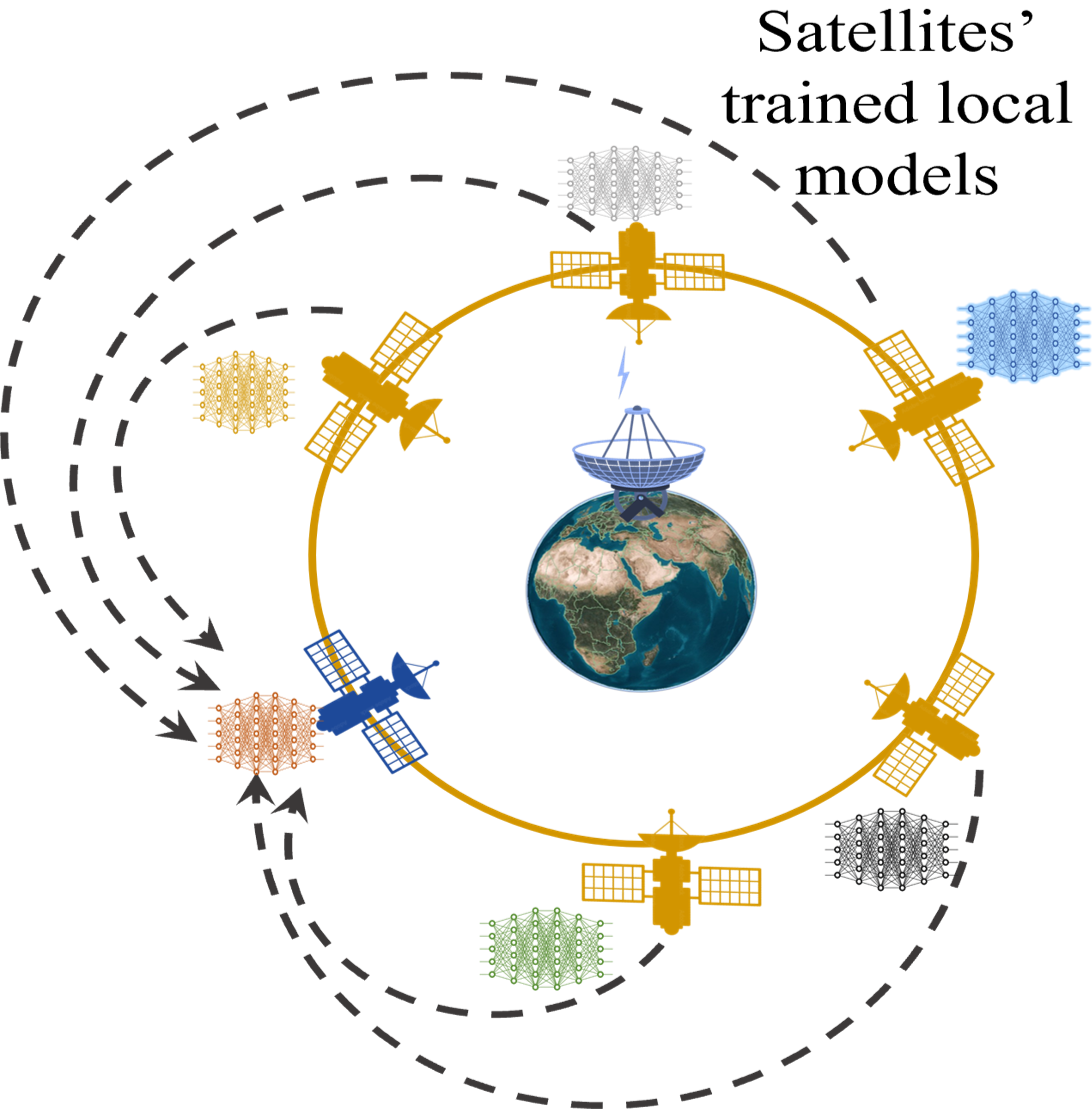}}
    \caption{Illustration of model propagation scheme.}
    \label{fig:propagation}
    \vspace{-0.4cm}
\end{figure*}

The model propagation scheme  augments the conventional ``vertical'' star topology between the FL server and LEO satellites with ``horizontal'' intra-plane ISL pathways among satellites to improve efficiency. \fref{fig:propagation} depicts this scheme. On any orbit $l$, once a visible satellite $k$ receives the global model $w^{t}$ from the GS, the satellite $k$ will send the model to $k$'s neighbors via intra-plane ISL\footnote{A LEO satellite has four antennas, two on the pitch axis and two on the roll axis for inter-plane and intra-plane communications, respectively.\vspace*{14mm}}, and each of these neighbors will further forward this model to its next-hop neighbor. In the meantime, each satellite on $l$ also starts to train its local copy $w^{t}$ upon receiving it (so there will be multiple concurrent training processes). During model propagation on an orbit, a satellite might receive $w^{t}$ twice from its neighbors (e.g., when more than one satellite is visible to the GS), which does not matter since it can simply drop the duplicate.
%To simplify the description, we describe model propagation for one orbit only, which applies to the other orbits as well.

After training, \fref{fig:propagation}c depicts the process of relaying the trained local models (represented by different colors) to a ``sink'' satellite via one or multiple ISL hops. The sink is the only one on each orbit that is responsible for {\em partial aggregation of models} and {\em sending the partial model back to the GS}, and is selected by a scheduler running on each satellite (so every satellite knows who is the sink) and explained in the next subsection. Upon having collected all the trained models from the same orbit $l$, the sink will aggregate them to generate a {\em partial global model} $w_{\mathcal K_{l}}$ as follows: 
\begin{equation}
    w_{\mathcal K_{l}} = \sum_{k\in \mathcal K_{l}} \frac{m_{k}}{m_{\mathcal K_{l}}} w_{k}^{I},
    \quad \text{\small where } m_{\mathcal K_{l}} = \sum_{k\in \mathcal K_{l}} m_k.
    \end{equation}
After that, the sink satellite uploads $w_{\mathcal K_{l}}$ along with the satellites' data distribution (in terms of the number of class labels, which is piggybacked onto the model propagation process) to the GS.

In the conventional star topology, where each satellite individually communicates with the GS, the total time $T_{sum}$ required for all the satellites on an orbit $\mathcal{K}_{l}$ to exchange their models with the GS can be defined in two intervals as: $T_{sum} =$ 
\begin{equation}\label{eq:sum}
\Resize{8.325cm} {
\begin{cases}
\sum_{k=1}^{\mathcal{K}_{l}} \big(2 t_{c}(k) + t_{wait}(k) +t_{train}(k)\big), ~\text{if~} t_{train}(k)< t_{visible}(k)\\ %\text{or}
\sum_{k=1}^{\mathcal{K}_{l}} \big(2 t_{c}(k) + 2t_{wait}(k) +t_{train}(k)\big), ~\text{if~} t_{train}(k)\geq t_{visible}(k)
\end{cases}
}
\end{equation}
where $t_{wait}(k)$ is the time that the GS must wait before the \textit{k-th} satellite enters its visible zone and $t_{visible}(k)$ is the visibility period of a satellite $k$ with the GS. The below $t_{train}(k)$ is the time required for a satellite $k$ to train a local model
\begin{equation}
     t_{train}(k)=  \frac{I (n_k b_k c_{k} )}{f_{k}}, 
\end{equation}
where $n_k$ is the number of mini-batches, $b_k$ is the size of each mini-batch, $c_{k}$ is the average number of processing cycles needed for training a data sample on $k$ satellite, and $f_{k}$ is the CPU/GPU frequency. In \eqref{eq:sum}, we consider two possible scenarios of calculating $T_{sum}$: (1) a satellite is able to complete its local model training within its visible period, or (2) a satellite requires an {\em additional round of waiting} for becoming visible again (to upload its trained model to the GS). %For this scenario, however, we only count the training time once (for the last satellite $K_l$) because others train their models in parallel during their invisible periods.}
%(satellites with a low inclination angle)
%{In the case of satellites with a high inclination angle (shorter visibility time), the GS first transmits $w^t$ to all satellites when they enter its visible zone. Then, satellites train their local models in parallel (count $t_{train}$ once for the last satellite $K_l$) during their invisible periods. As soon as they become visible again to the GS, they upload their models to the GS.}%  (for the last satellite $K_l$) because others' are in parallel with the waiting time $t_{wait}$}.
The above represents a largely sequential procedure (note the accumulation of $t_{wait}$ and $t_c$). In contrast, by introducing model propagation, $t_{wait}$ and $t_c$ will occur to the sink satellite {\em only}. In other words, we can significantly reduce the total time to\vspace{-1mm}
%To accomplish this, the selected satellites generate partial global models that represent the overall local models per each orbit, which will then serve as teacher models. 
\begin{equation}
     T_{sum}^* = 2 t_{c} + t_{wait}^* +t_{train}(\mathcal K_l), 
     \label{opt}
\end{equation}\enlargethispage{-3.5\baselineskip}
where $t_{wait}^*$ is the time waiting for the sink node to be visible to the GS, and $t_{train}(\mathcal K_l)= \max_{k\in \mathcal K_{l}} t_{train}(k)$ is the parallel training time of all satellite models on orbit $l$. Note that we take advantage of the fact that the process of relaying the model to the sink happens in parallel with $t_{wait}^*$. In fact, this relay is very fast on its own because ISL typically uses FSO links, which support a much higher data rate (Gbps to Tbps) than RF links. However, in our simulation, we {\em forgo this benefit} of using FSO links so that {\em all our performance advantages over baseline approaches come from our architecture and algorithm design, rather than being mixed with hardware benefits}.

\subsection{Distributed Scheduling} \label{sec:schedule}
\begin{figure}[ht!]
\vspace{-0.2cm}
         \centering{ \includegraphics[width=0.7\linewidth]{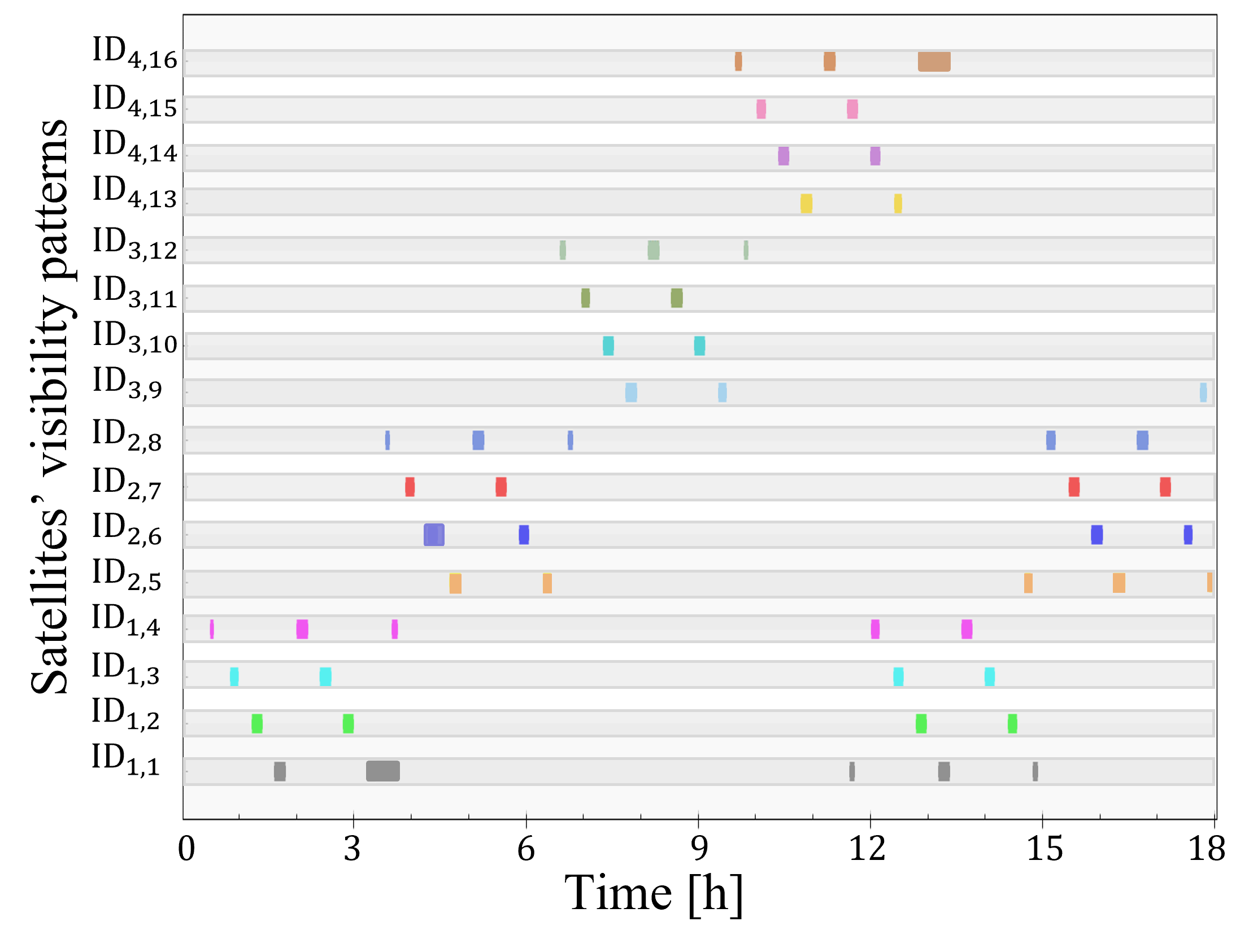}}
         \caption{A visiting pattern over 18 hours of a Walker-delta constellation consisting of 16 satellites that communicate with a GS in Rolla, MO, USA ($ID_{l,k}$ is the {\em k-th} satellite in an orbit $l$).}
\label{Scheduling}\vspace{-0.2cm}
\end{figure}

We propose a distributed scheduling algorithm to select an optimal ``sink'' satellite on each orbit to generate a partial global model, and optimally schedule communication between each sink satellite and the GS by exploiting the predictability of the satellite orbiting pattern. \fref{Scheduling} shows a real visiting pattern of 16 satellites evenly distributed over four orbits (i.e., 4 on each orbit) communicating with a GS. Although these satellites travel at the same speed $v_{l}$ with the same orbital period $T_{l}$ and are equally spaced on each orbit, they have different and irregular visibility periods and varying numbers of visits to the same GS; even each satellite's own visits are not equally spaced and have different durations. This is because 1) the trajectory of the Earth differs from that of a satellite (which has an inclination angle between 0--90$^o$) and 2) LEO satellites travel at very high speeds ($\approx$ 7.8 km/s) in comparison to Earth's rotational speed ($\approx$ 0.45 km/s).\enlargethispage{-3.5\baselineskip}

To tackle the high irregularity of the satellite-GS visiting pattern, we propose a distributed scheduling algorithm that selects a {\em sink satellite} for each orbit via prediction of its visibility period (and the sink changes over time). This scheduler runs at each satellite once the satellite completes its local training of model $w^t$, in order to determine which satellite on the same orbit will be the first next satellite to visit the GS for a visiting period longer than the minimum interval of sending all $|\mathcal{K}_l|$ models to the GS. This satellite is referred to as the sink satellite (one per orbit). We predict the satellite visibility using a method introduced in \cite{ali1999predicting}. Once the sink is determined, each satellite sends its locally trained model to the sink using our propagation algorithm discussed in Section \ref{propagate}. {The sink $k_s$ then bags all these local models into a {\em partial global model} and uploads it to the GS when $k_s$ first becomes visible to the GS. In other words, our distributed scheduling algorithm leverages the predictable satellite-GS visibility despite its sporadic nature. 

%\red{in below, you use uplink and downlink to represent model upload/download directions, but usually up/downlink refers to the reverse direction, so such terms will confuse reviewers.}\blue{Do you mean by the reverse direction that the downlink will be from clients to the server and vice versa? I tried to reverse this paragraph based on my understanding, so could you double-check it again?}\red{yes, downlink: sat to GS, uplink: GS to sat}\blue{That is correct. I was confused as I thought the GS during the downlink is downloading the Sats data.]}

Suppose there are $N$ resource blocks (RBs) of bandwidth $B^{D}$ each, for exchanging models between satellites and the GS. To ensure the smooth operation of the iterative learning process, the GS needs to allocate the entire bandwidth $B$ (i.e., $B=NB^{D}$) to all visible satellites for broadcasting the global model via the uplink. On the other hand, sink satellites from multiple orbits will compete for an RB to send their partial global models to the GS via the downlink (using a multiple access protocol such as ALOHA). As a result, SNRs for uplink and downlink channels between a satellite $k$ and the GS can be expressed as\vspace{-2mm}
\begin{align}
\Resize{7.8cm} 
{SNR(GS)^{U}= P_t(GS)+G_{k}+G_{GS}-L_{(k,GS)}-K_{B}TB     ~~~(dB)}\\
\Resize{7.8cm} 
{SNR(k)^{D}= P_t(k)+G_{k}+G_{GS}-L_{(k,GS)}-K_{B}TB^{D}  ~~~(dB)}
\vspace{-3mm}
\end{align}\enlargethispage{-3.5\baselineskip}
Thus, the overall communication latency given in (\ref{eqny}) can be reformulated as\vspace{-2mm}
\begin{align}
 \textstyle t_{c}^{U} \approx \frac{z{|\bf{\mathcal{N}|}}}{B \log_{2} (1+SNR(GS)^{U})}+\frac{\|k,GS\|_{2}}{c}+t_{k}+ t_{GS}\label{download_time}\\
\textstyle t_{c}^{D} \approx \frac{z{|\bf{\mathcal{N}|}}}{B^{D} \log_{2} (1+SNR(k)^{D})}+\frac{\|k,GS\|_{2}}{c}+t_{k}+ t_{GS} \label{upload_time}
\end{align}
Accordingly, the total time required by each sink satellite to exchange models with a GS can be determined by substituting (\ref{download_time}, \ref{upload_time}) in (\ref{opt}), yielding\vspace{-2mm}
\begin{equation} 
     T_{sum}^* = t_{c}^{U}+t_{c}^{D}+ t_{wait}^*+t_{train}(\mathcal K_l), 
     \label{opt2}\vspace{-3mm}
\end{equation}
In addition, we denote the start time and end time of the {\em access window} $AW(k,GS)$ between a satellite $k$ and the GS by $t_{start}(k,GS)$ and $t_{end}(k,GS)$, respectively. The start time is the time when a satellite $k$ enters the visibility zone of the GS, and the end time is the time when $k$ leaves the visibility zone. Consider all the $K$ satellites,

\vspace{-3mm}
\begin{equation}
\begin{aligned}
\Resize{7.97cm}{ \textstyle {T_{start}} = \Big\{ \{ t_{start, (k=1,GS)}^{r}\}_{r\in \mathcal{R}_{1}},\{ t_{start, (k=2,GS)}^{r}\}_{r\in \mathcal{R}_{2}}}\\,\dots, \{ t_{start, (k=K,GS)}^{r}\}_{r\in \mathcal{R}_{K}} \Big\}\\
\textstyle {T_{end}} = \Big\{ \{ t_{end, (k=1,GS)}^{r}\}_{r\in \mathcal{R}_{1}},\{ t_{end, (k=2,GS)}^{r}\}_{r\in \mathcal{R}_{2}}\\,\dots, \{t_{end, (k=K,GS)}^{r}\}_{r\in \mathcal{R}_{K}} \Big\},
\end{aligned}
\end{equation}
where $t_{start, (k,GS)}^{r}$ and $t_{end, (k,GS)}^{r}$ refer to $t_{start, (k,GS)}$ and $t_{end, (k,GS)}$ for the ${r}$-th visit, respectively, i.e., $k$ enters and leaves the GS's visibility zone for the $r$-th time. $\mathcal{R}_k=\{1,2,.., R_k\}$ is the index set of all the visits of satellite $k$. Thus, the AW can be written as\vspace*{-2mm} %\red{$r$ is also unclear, do your $\{ t_{start, k=1}^{r}\}_{r\in \mathcal{R}_{1}}$ mean a set of all the possible $r=1,2,...$? }
\enlargethispage{-3.5\baselineskip}
\begin{equation} 
\begin{aligned}
     AW(k,GS) =& \bigcup_{k\in\mathcal{K}} \Big\{  \big[ T_{start}, T_{end}\big] \Big\} \\
     =& \Big\{ \Resize{5.5cm}{\big \{ \big[ t_{start(k=1,GS)}^{r}, t_{end (k=1,GS)}^{r} \big]\big\}_{r\in \mathcal{R}_{1}}}\\,\dots, & \big\{\Resize{5.5cm}{ \big[ t_{start (k=K,GS)}^{r}, t_{end (k=K,GS)}^{r} \big]_{r\in \mathcal{R}_{K}}\big\} \Bigl\}}
\end{aligned}
\end{equation}
Suppose that each ISL hop $h$ between any two adjacent satellites is allocated one RB with bandwidth $B^{h}$. %This RB can be shared with multiple ISL hops between two distant satellites via time-division. $\mathcal H=\{1,2,\dots, H\}$
Accordingly, the time required for exchanging a model between two adjacent satellites $(k,k+1)$ via a single hop $h$ is \vspace{-3mm}
\begin{equation}
    t_{h}(k,k+1)=\frac{z{|\bf{\mathcal{N}|}}}{B^{h} \beta_{h}}
\end{equation}
where $\beta_{h}$ is the spectral efficiency of the communication link. 

Consider a set of candidate sink satellites $\mathcal{C}_l=\{1, 2,..., C_{l}\}$ on an orbit $l$, each of which satisfies the requirements on training time $t_{train}(\mathcal K_l)$ and communication latency  $t_c^U$ and $t_c^D$ (see eqn. \ref{opt}, \ref{download_time}, \ref{upload_time}). On each orbit, to determine the optimal sink satellite $c_{opt}$ for exchanging models with the GS, we formulate an optimization problem that minimizes a sum that involves (1) the waiting time $t{_{\text{wait}}^{*}}$ for the sink satellite to enter the GS's visible zone, and (2) the model relaying time ${t_{h}^*(k,\mathcal{C}_l)}$ for each other satellite $k$ to reach the sink, where
\begin{equation} 
      t_{h}^* (i, j)=  \max_{i\in \mathcal K_{l}:i\neq j,\forall j\in \mathcal C_{l} } \frac{h ~ z{|\bf{\mathcal{N}|}}}{B^{h} \beta_{h}} ,~~~~~ h=1,2,\dots, H 
\end{equation}
Therefore, the optimal sink satellite for each orbit should minimize the overall latency as 
\begin{equation} 
\begin{aligned}
     T_{sum}^* =  \min\bigl\{t_{c}^{D}+t_{c}^{U}+t_{wait}^*+t_{train}(\mathcal K_l)+t_{h}^* \bigl\}    
\end{aligned}
\end{equation}
Given this, the scheduler selects the optimal sink satellite $c_{opt}$ to be the one whose AW satisfies the time constraint $AW(c_{opt},GS)\geq T_{sum}^*$. In the case of more than one, the scheduler selects the one that will visit the GS the first to be the  optimal sink.

%\red{your math will turn off every reviewer, but fortunately most people don't read math so that's why in previous papers they seldom commented on anything wrong of your math. anyway, now we don't have time to rewrite all the math. just fix the errors I pointed out. finally, I feel in the entire paper, many math is unnecessary, but again, no time now, just keep them.}\blue{ You are right, I noticed some math issues when revising the paper this time. But I think if we correct those issues that you pointed out, then the paper should be fine.}

\vspace*{-1mm}

\section{Performance Evaluation}

%We evaluate SatShot under a variety of different settings and compare it with state-of-the-art methods.
%\blue{I have to consider simulation results when images download to a GS and generate a centralized model. In addition, I have to use realistic datasets and LEO configuration (geospatial dataset, torcghgeo, deepGlobe road  extraction dataset, }
\vspace*{-1mm}
\subsection{Simulation setup}

\textbf{LEO Constellation.} We consider a Walker-delta constellation $\mathcal K$ consisting of 40 LEO satellites distributed evenly on five orbits. Each orbit $l$ is located at an altitude $h_{l}$ of 1500 km above the Earth's surface with an inclination angle of 80 degrees. A GS is located in Rolla, MO, USA (but can be anywhere on the Earth) with a minimum elevation angle of 10 degrees. Communication link parameters as described in \sref{Com_link} are assigned according to the values in Table \ref{Parameter} (upper part). We use the Systems Tool Kit (STK), a software tool for analyzing satellite constellations, to compute the visiting pattern of LEO satellites with regard to the GS. To obtain each set of results, we run each simulation over a period of three days.
\begin{table}[t!]
\vspace{0.2cm}
\setlength{\tabcolsep}{0.75em}
%\centering\arraybackslash
\centering
\renewcommand{\arraystretch}{0.9}
\caption{Simulation Parameters (Comms: upper; Learning: lower)}
\label{Parameter}
%\scriptsize
% \begin{tabular}{|p{5.3cm}|p{1.3cm}|} 
\begin{tabular}{|l|l|}
 \hline
  \centering \textbf{Parameters} & \textbf{Values} \\
   \hline  
   \hline \hline
  Transmission power (satellite \& GS) $P_{t}$& 40 dBm \\
  \hline
  Antenna gain of (satellite \& GS) $G_k , G_{GS}$& 6.98 dBi\\
  \hline
 Carrier frequency $f$ & 2.4 GHz \\
  \hline 
 Noise temperature $T$& 354.81 K \\
  \hline 
 Transmission data rate $R$ & 16 Mb/sec   \\ 
 \hline 
 \hline 
 Average number of processing cycles $c_k$ & $10^3$  \cite{zeng2020federated}  \\ 
 \hline 
CPU/GPU frequency $f_k$ & $10^9$ \cite{zeng2020federated}\\
  \hline 
   Number of local training epochs $I$ & 100 \\
  \hline 
   Learning rate $\eta$ & 0.001 \\
  \hline 
   Mini-batch size $b_k$& 32\\
\hline 
\end{tabular}\vspace{-3mm}
\end{table}

\textbf{Baselines.} We compare FedLEO with state-of-the-art (SOTA) approaches proposed recently and reviewed in \sref{sec:releated}: {\em sync} approaches including FedAvg~\cite{mcmahan2017communication}, FedISL~\cite{razmi} and FedHAP~\cite{happaper}, and {\em async} approaches including FedAsync~\cite{xie2020asynchronous}, FedSpace~\cite{so2022fedspace}, FedSat~\cite{razmi2022ground}, FedSatSched \cite{razmischeduling}, and AsyncFLEO \cite{mAsyFLEO}.\vspace{-1mm}
\begin{table}[ht!]
\setlength{\tabcolsep}{0.4em}
%\centering\arraybackslash
\centering
\renewcommand{\arraystretch}{1}
\caption{FedLEO vs. SOTA FL under non-IID.} 
\label{table1}
%\scriptsize
\resizebox{3.3in}{!}{
 \begin{tabular}{|p{2.1cm}|p{0.8cm} | p{1.2cm} | p{0.9cm} | p{3.7cm} |}
 \hline
 \centering FL &\multicolumn{2}{c |} {Accuracy (\%)} & \centering Conv & Remark \\
 \cline{2-3}
\centering Approaches &\rmfamily MNIST& CIFAR-10&time (h)& \\
 \hline 
FedAvg \cite{mcmahan2017communication} & 85.84& 63.26& 48& GS at any location \\
 \hline
FedISL \newline ({\em ideal setup}) \cite{razmi} &  83.67& 75.31&4 & GS at NP; MEO above Equator\\
 \hline
FedISL \cite{razmi} &  63.45& 54.33&72& GS at any location \\
 \hline
FedHAP \cite{happaper} &  83.94&80.89 & 40& GS at any location \\
 \hline
FedHAP \cite{happaper} &80.45 (89.83)&77.67 (84.31) &5 \newline(30)& using two HAPs as server not a GS\\
 \hline
  {FedAsync} \cite{xie2020asynchronous} & 81.63& 59.18& 48& GS at any location \\
   \hline
  FedSpace \cite{so2022fedspace}& 53.12&40.69&72 & Sats need to upload raw data to the GS \\ 
   \hline
  FedSat \newline({\em ideal setup}) \cite{razmi2022ground} &  88.65 &69.19 & 24&Sats visit GS at regular interval \\
   \hline
  {FedSatSched} \cite{razmischeduling} & 76.32& 70.95& 48& GS at any location \\
    \hline
  {AsyncFLEO} \cite{mAsyFLEO} & 80.62& 77.23& 6& GS at any location (ignores sink's visible period) \\
    \hline
    \rowcolor{gray!30}
    \textbf{FedLEO} &  \textbf{89.37}& \textbf{82.13}& \textbf{15}& GS at any location \\
 \hline 
\end{tabular}
}
\end{table}\vspace{-1mm}

\textbf{Dataset and ML models.} To compare FedLEO with the SOTA FL-Satcom approaches, we use the same datasets they used (MNIST and CIFAR-10). %\cite{deng2012mnist}%\cite{CIFAR-10}%In the MNIST dataset, there are 70,000 grayscale images of handwritten numbers of size 28$\times$28 pixels, while CIFAR-10 contains 60,000 colored images of size 32$\times$32 pixels (tiny images of animals and vehicles). 
But in addition, we also use a real satellite dataset, DeepGlobe Road Extraction \cite{DeepGlobe18}. This DeepGlobe dataset contains 6,226 colored satellite images of size 1024$\times$1024 where each pixel represents a resolution of 50 cm, and these images have been annotated for road extraction. We augment the DeepGlobe dataset via flipping, rotating, varying lighting conditions, etc. to generate more training data. For the ML model, we use a deep CNN for MNIST and CIFAR-10, and a U-Net model for DeepGlobe. We consider both IID and non-IID data distributions among satellites (except for DeepGlobe which is non-IID by nature). In the IID setting, training images are randomly shuffled and equally distributed across all satellites, with each satellite having all 10 classes. In the non-IID setting, satellites in two orbits are trained on 4 classes, while satellites in the remaining three orbits are trained on the other 6 classes. Table \ref{Parameter} (the lower part) summarizes the training hyperparameters.\vspace{-0.20cm}

\subsection{Results}
\enlargethispage{-3.5\baselineskip}
{\bf Comparison with SOTA}. Table \ref{table1} shows a comparison of FedLEO's performance with SOTA using the MNIST and CIFAR-10 datasets in the non-IID setting (results for IID have a similar trend and are omitted due to space constraints). According to the results for the MNIST dataset, FedLEO achieves an accuracy of 89.37\% within only 15 hours (including  the waiting time for sink satellites to enter the visibility zone of GS). With the ideal assumption of FedISL \cite{razmi} where the GS is located at the NP or is an MEO satellite orbiting above the Equator, FedISL converges within 4 hours with an accuracy of 83.67\%. However, when we omit this assumption (GS at any location), the FedISL's accuracy is reduced to 63.45\% within  72 hours of training. The accuracy of FedSat \cite{razmi2022ground} and FedHAP \cite{happaper} are marginally close to FedLEO, however, their FL convergence processes are considerably slower. In addition, FedSat presumes a very similar setup to FedISL, limiting its application, and FedHAP requires extra hardware (HAP) to achieve better results. When FedSatSched \cite{razmischeduling} removes FedSat's ideal assumption, its accuracy drops to 76.32\%, and the convergence time doubles. AsyncFLEO \cite{mAsyFLEO} converges in only 6 hours with an accuracy of 80.62\%; however, it does not consider whether the sink node will have sufficient time to train and exchange models with the GS.
%\teal{When using the MNIST dataset in the IID setting, most of these FL approaches achieve  similar levels of accuracy more quickly (5-24 hours less). In FedLEO, for example, 89.67\% is reached after 10 hours only, however, the IID setting does not reflect an actual scenario.}
\begin{figure}[ht!]
\centering
\vspace*{-3mm}
          \subfloat[Satellite image \& ground-truth label (extracted roads).]
         {\includegraphics[width=0.35\textwidth]{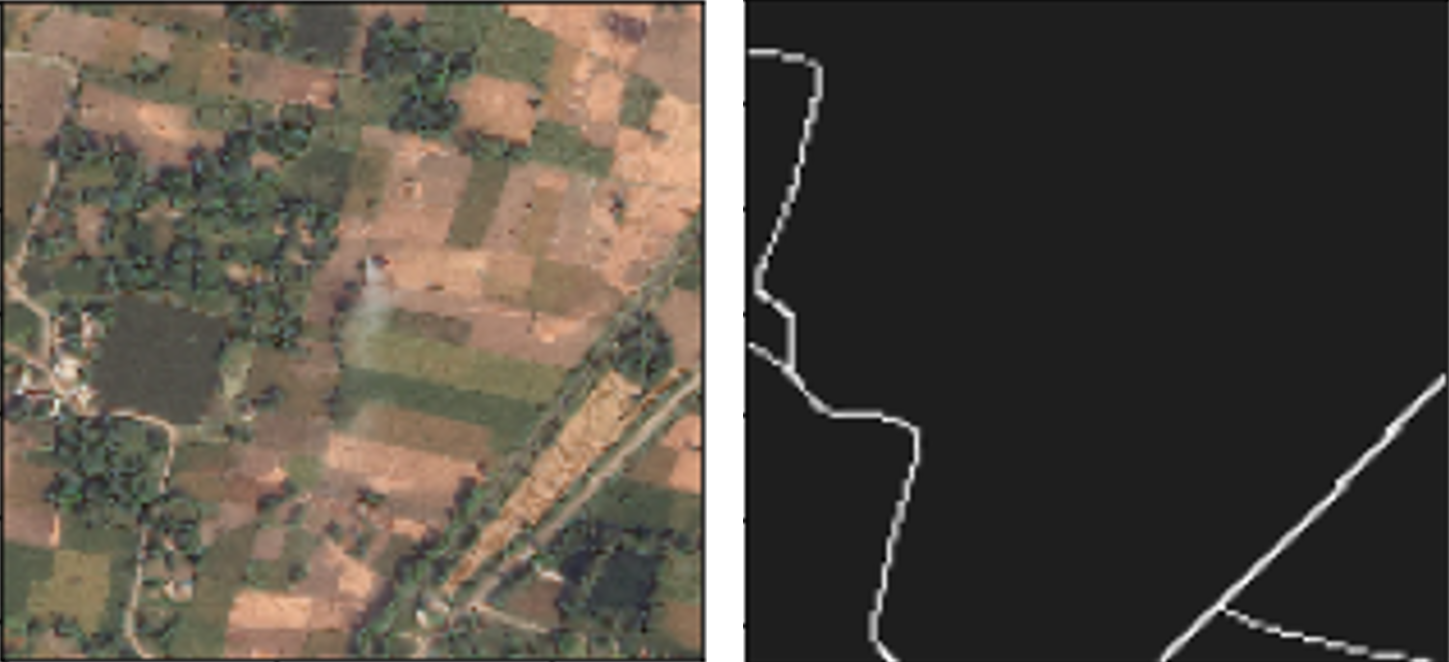}}
     \hfill
          \subfloat[8 hours vs. 22 hours road prediction.]
         {\includegraphics[width=0.35\textwidth]{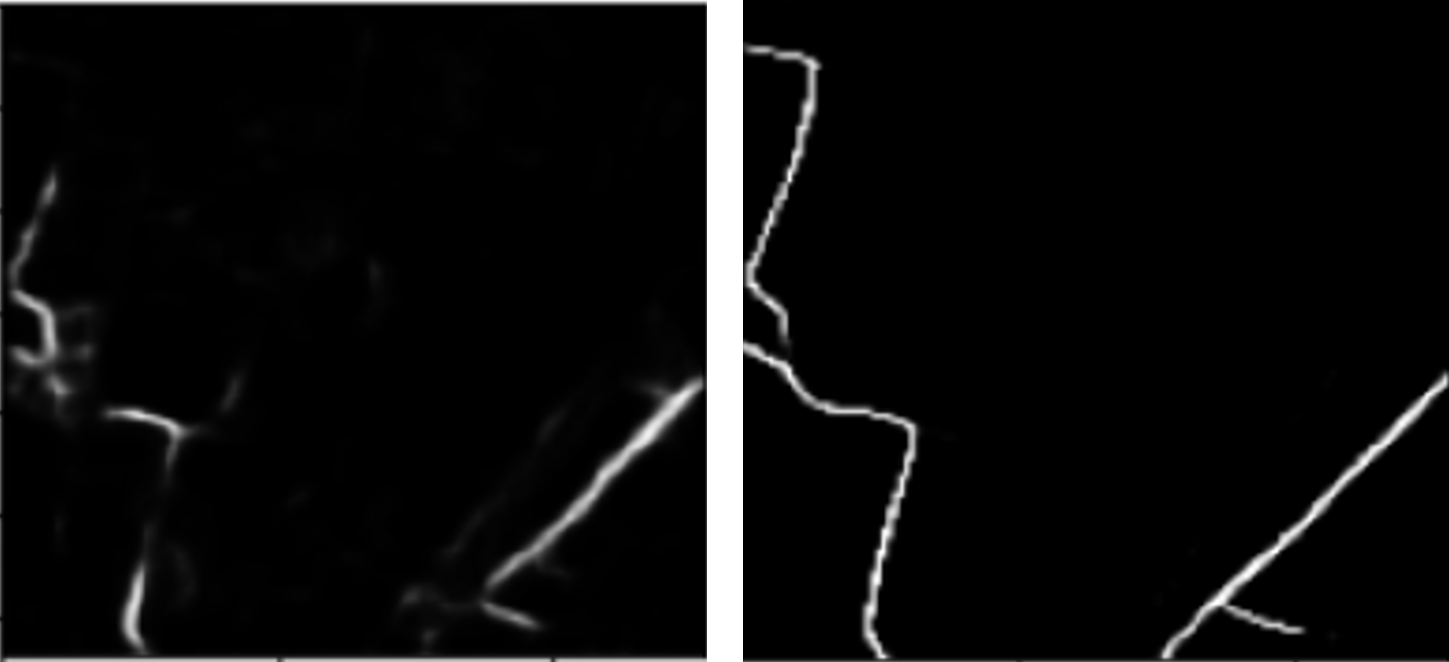}}
\caption{Sample of FedLEO's performance on the DeepGlobe dataset.}
\label{fig:acc_time_round}
\vspace*{-4mm}
\end{figure}
\begin{figure}[ht!]
\centering \vspace{-1mm}
 \includegraphics[width=0.3\textwidth]{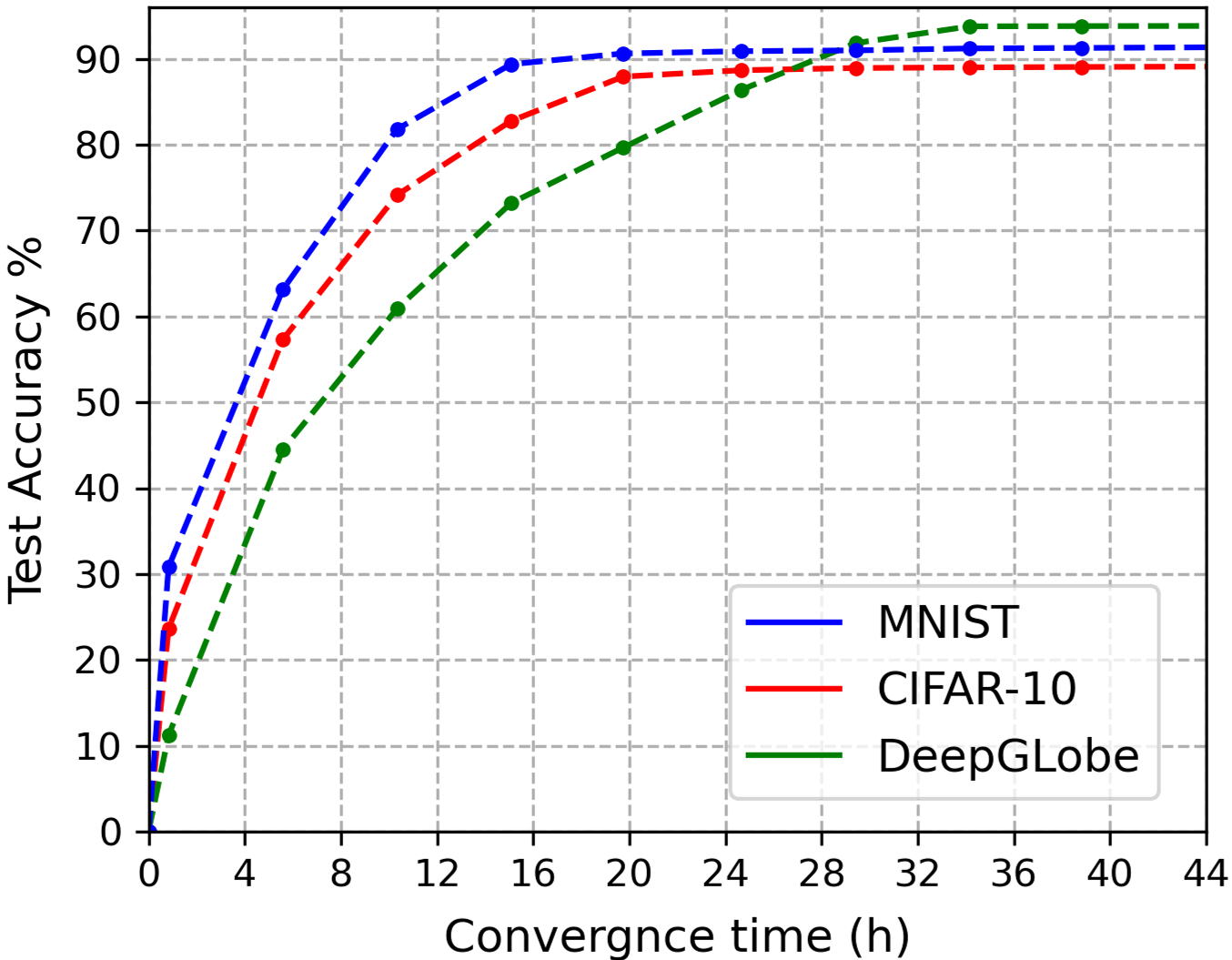}
\caption{FedLEO's accuracy vs. convergence on all datasets.}
\label{FedLEO_data}\vspace{-3mm}
\end{figure}
When the CIFAR-10 dataset is used, all the FL approaches  fail to achieve the same accuracy on the MNIST dataset within the same convergence timeframe (saliently for FedAvg \cite{mcmahan2017communication}, FedAsync \cite{xie2020asynchronous} and FedSat \cite{razmi2022ground}), while FedLEO still maintains the highest accuracy 82.13\% within only 15 hours, proving its robustness. 
\enlargethispage{-3.5\baselineskip}
{\bf Evaluating FedLEO on DeepGlobe.} DeepGlobe is a more sophisticated and realistic dataset for LEO satellites. Fig. \ref{fig:acc_time_round} shows how FedLEO recognizes the roads from a DeepGlobe image at two different times of convergence. The accuracy was 52.37\% in 8 hours and then increased to 82.77\% after 22 hours. %Although the convergence time is increased to 32 hours, the accuracy is also increased by 4\% over MNIST (which is simple compared to DeepGlobe). 

Fig.~\ref{FedLEO_data} compares the accuracy vs. convergence time of FedLEO. It can be seen that FedLEO can achieve similar or even better accuracy on DeepGlobe than on MNIST and CIFAR-10, but requires extra time due to the complexity of the dataset. Other SOTA FL approaches take much longer convergence time with lower accuracy under the same setting. %This proves the robustness and effectiveness of FedLEO in generating an FL global model with at least {\em fivefold increase} in speed compared to SOTA. \blue{Should I add anything about our future work? Does a figure need to be included or will that be sufficient to convince the reviewer? If not, I will include a figure. }
\vspace{-1mm}
\section{Conclusion} \label{sec:conc}
We have proposed FedLEO, an efficient FL framework designed for LEO satellite constellations. FedLEO tackles the challenging problem of highly intermittent and irregular patterns in which LEO satellites visit GSs. FedLEO consists of a model propagation algorithm and a distributed scheduling algorithm to achieve the goal. In our quantitative study with several SOTA benchmarks, we find that FedLEO drastically reduces FL training/convergence time; in the meantime, it achieves an accuracy of 89.37\% even under the tough non-IID setting, outperforming the benchmarks by a wide margin. We also evaluate FedLEO on a real satellite dataset (DeepGlobe), which no prior work in this area has done before.

\bibliographystyle{IEEEtran}
\small{\vspace*{-4mm}
\bibliography{references.bib}
}
\end{document}